\RequirePackage{fix-cm}

\documentclass[smallextended]{svjour3}       
\smartqed  
\usepackage{graphicx, longtable, rotating, soul, array, multirow, colortbl}
\usepackage[numbers]{natbib}
\usepackage{hyperref}
\usepackage{float}
\usepackage{caption, tabularx, rotating}
\renewcommand{\figurename}{\bf Fig.}
\begin{document}

\title{ShEMO $-$ A Large-Scale Validated Database for Persian Speech Emotion Detection 
}

\author{Omid Mohamad Nezami \and
		Paria Jamshid Lou   \and
        Mansoureh Karami 
}

\institute{O. Mohamad Nezami \at
              Islamic Azad University of Bijar, Bijar, Iran \\
              \email{odnzmi@gmail.com}           
           \and
           P. Jamshid Lou \at
           Sharif University of Technology, Tehran, Iran\\
           \email{paria.jamshidlou@gmail.com}  
           \and
           M. Karami \at
           Sharif University of Technology, Tehran, Iran
}
\titlerunning {A Large-Scale Validated Database for Persian Speech Emotion Detection} 
\date{Received: date / Accepted: date}

\maketitle

\begin{abstract}
This paper introduces a large-scale, validated database for Persian called \emph{Sharif Emotional Speech Database (ShEMO)}. The database includes 3000 semi-natural utterances, equivalent to 3 hours and 25 minutes of speech data extracted from online radio plays. The ShEMO covers speech samples of 87 native-Persian speakers for five basic emotions including \emph{anger}, \emph{fear}, \emph{happiness}, \emph{sadness} and \emph{surprise}, as well as neutral state. Twelve annotators label the underlying emotional state of utterances and majority voting is used to decide on the final labels. According to the kappa measure, 
the inter-annotator agreement is $64\%$ which is interpreted as ``substantial agreement''. We also present benchmark results based on common classification methods in speech emotion detection task. According to the experiments, support vector machine achieves the best results for both gender-independent ($58.2\%$) and gender-dependent models (female=$59.4\%$, male=$57.6\%$). The ShEMO is available\footnote{Data available at: \url{https://github.com/pariajm/ShEMO}} for academic purposes free of charge to provide a baseline for further research on Persian emotional speech.

\end{abstract}

\section{Introduction}
\label{intro}
Speech emotion detection systems aim at recognizing the underlying affective state of speakers from their speech signals. These systems have a wide range of applications from human-machine interactions to auto-supervision and control of safety systems~\cite{hu:10}. For example, negative/positive experience of customers can be automatically detected in remote call centres to evaluate company services or attitude of staff towards customers~\cite{batliner:03}. These systems can also be used in health domains to monitor and detect the early signs of a depression episode~\cite{dickerson:11} or help autistic children learn how to recognize more subtle social cues~\cite{hen:16}. Another application is crime detection where the psychological state of criminal suspects (i.e. whether or not they are lying) is discovered~\cite{cowie:01}. They are also useful for in-car board systems where information of the driver's emotion is extracted to increase their safety~\cite{schul:04}. Furthermore, identifying the affective state of students in academic environments can help teachers or intelligent virtual agents to provide students with proper responses and improve teaching quality accordingly~\cite{kort:01}.

An important issue that should be considered before developing any speech emotion detection systems is the quality of database. In fact, the performance of these systems (like any statistical models) depends on the quality of training data~\cite{busso:13}. On the other hand, there is usually a lack of decent benchmark emotional speech database for non-English languages such as Persian. As some studies~\cite{fur:87, fer:15, sagha:16} show, the relation between linguistic content and emotion is language dependent, so generalization from one language to another language is often difficult. That's why speech emotion detection systems are usually developed language-dependently.

A few studies have explored Persian speech emotion detection and introduced emotional databases~\cite{esm:13, sav:15, kesh:15, moos:07, ghar:08, man:09, ham:12}. Persian Emotional Speech Database (Persian ESD)~\cite{kesh:15} and Sahand Emotional Speech Database (SES)~\cite{seda:08} are two important datasets in Persian. Although its validity has been evaluated by a group of native speakers, Persian ESD covers emotional speech of only two speakers, which is not large enough for developing a robust system. SES covers the emotional speech of 10 speakers and is larger than Persian ESD but its reliability is relatively low according to the results of perception test. 

In this paper, we present a large-scale validated dataset for Persian called \emph{Sharif Emotional Speech Database (ShEMO)}. The ShEMO is a semi-natural dataset which contains emotional (as well as neutral) speech samples of various Persian speakers. In addition to collecting the dataset, we benchmark the performance of standard classifiers on this dataset and compare the results to other languages to provide a baseline for further research. To the authors' best knowledge, this is the first systematic effort towards creating a large validated emotional speech dataset and corresponding benchmark results for Persian. The ShEMO database will be publicly available to facilitate research on Persian emotional speech\footnote{Upon publishing this paper, we release our database for academic purposes.}. 

The remainder of this paper is organized as follows. In Section~\ref{sec:1}, we review different types of emotional speech databases and explain the efforts made so far for designing/collecting a database for Persian. We introduce the ShEMO database and describe the process of data collection, annotation and validation in Section \ref{sec:2}. We also discuss the baseline performance of standard classification methods on the ShEMO dataset and compare it to other databases in Persian, German and English. Finally in Section~\ref{sec:4}, we  summarize our analysis and suggest future directions with our dataset.

\section{Related Work}
\label{sec:1}
Due to the vast amount of literature on emotional speech in general, this section will focus on reviewing different types of emotional speech database and the efforts made so far for data collection and validation for Persian language.

\subsection{Types of Emotional Speech Database}
Emotional speech databases can be categorized in terms of \emph{naturalness}, \emph{emotion}, \emph{speaker}, \emph{language}, \emph{distribution} and so forth~\cite{ayadi:11}. Naturalness is one of the most important factors to be considered when designing or collecting a database. Based on the degree of naturalness, databases can be divided into three types of \emph{natural}, \emph{semi-natural} and \emph{simulated}~\cite{ayadi:11}. In natural databases, speech data is collected from real-life situations to guarantee that the underlying emotions of utterances are naturally conveyed. Such databases are rarely used for research purposes due to the legal and ethical issues accompanied with data collection. To avoid the difficulty, most natural databases are built by recording the emotional speech of some volunteer or recruited participants whose emotions have been naturally evoked by a method. For instance, a natural database may cover speech samples of non-professional actors discussing emotional events of their lives. Belfast Induced Natural Emotion Database~\cite{doug:00} is an example where individuals' discussion about emotive subjects and interactions among the audience in television shows induce emotional speech. Computer games can also be used to naturally elicit emotional speech since players usually react positively or negatively towards winning or losing a game~\cite{john:05}. Another technique is \emph{Wizard-of-Oz} scenario~\cite{bat:00} where a human, so-called \emph{Wizard}, simulates a dialogue system to interact with the users in such a way that they believe they are speaking to a machine. For instance, FAU Aibo Emotion Corpus~\cite{stei:09} contains the spontaneous emotional speech of children talking to a dog-like robot. The Aibo robot is controlled by a human wizard to show obedient and disobedient behaviors so that the emotional reactions of children can be induced. 

Another type of emotional speech database is semi-natural which is built using either scenario-based approach or acting-based one. In the scenario-based approach~\cite{sch:91}, the affective state of speakers is first evoked by a method. For instance, speakers recall some memories or read given sentences describing a scenario to get emotional. Then, they are asked to read a pre-written text in a particular emotion which aligns their provoked affective state. Persian Emotional Speech Database~\cite{kesh:15} is an instance of this type. In the acting-based approach, emotional utterances are extracted from movies or radio plays. To illustrate, Chinese Emotional Speech Database~\cite{yu:01} includes 721 utterances extracted from teleplays. Giannakopoulos~\cite{gia:09} also uses English movies to collect 1500 affective speech samples. 

Emotional speech databases can be simulated. For collecting this type of databases, scripted texts including isolated words or sentences are used. The prompt texts are usually semantically neutral and interpretable to any given emotion\footnote{The prompt excludes any emotional contents in order not to intervene the expression and perception of emotional states.}. For recording these databases, professional stage actors are recruited to express the pre-determined sentences or words in peculiar emotions. The utterances are usually recorded in acoustic studios with high quality microphones in order not to influence the spectral amplitude or phase characteristics of the speech signal. Berlin Database of Emotional Speech~\cite{bur:05} and Danish Emotional Speech Database~\cite{eng:97} are two examples of simulated data. The main disadvantage of simulated databases is that emotions are usually exaggerated and far from natural. To alleviate this problem, non-professional actors (such as academic students or employees) are hired to read the prompt.  

In addition to degree of naturalness, the theoretical framework of emotional speech databases can be different from each other. Two important theories of emotion include categorical and dimensional approach. According to the categorical approach, there are a small number of basic emotions which are recognized universally. Ekman~\cite{ekman:82} showed that \emph{anger}, \emph{fear}, \emph{surprise}, \emph{happiness}, \emph{disgust}, \emph{sadness} are six basic emotions which can be recognized universally~\cite{wol:13, mc:10, nic:11}. According to dimensional approach, affective states are not independent from one another, but they are systematically related so that they can be demonstrated as broad dimensions of experience. In this approach, emotions are represented as continuous numerical values on two main, inter-correlated dimensions of valence and arousal. The valence dimension shows how positive or negative the emotion is, ranging from unpleasant to pleasant feelings. The arousal dimension indicates how active or passive the emotion is, ranging from boredom to frantic excitement~\cite{rus:80, alv:97, lew:07}.

Emotional speech databases can also be differentiated in terms of speakers. In most cases, professional actors are recruited to read pre-written sentences in target emotions (e.g. Berlin Database of Emotional Speech). However, some databases use semi-professional actors (e.g. Danish Emotional Speech Database) or ordinary people (e.g. Sahand Emotional Speech Database) to avoid exaggerated emotion expression. Furthermore, the utterances of some datasets (e.g. Berlin Database of Emotional Speech) are uniformly distributed over emotions while the distribution of emotions in other datasets are unbalanced and may reveal their frequency in the real world (e.g. Chinese Emotional Speech Database). Another important factor is availability of databases. While the majority of emotional speech databases are private (e.g. MPEG-4~\cite{schul:05}), there are some datasets which are available for public use (e.g. FERMUS III~\cite{schul:02}, RAVDESS~\cite{liv:12}). 

\subsection{Persian Emotional Speech Databases}
In recent years, some efforts have been made to record and collect validated datasets in Persian. In this section, we will elaborate two important ones including Sahand Emotional Speech (SES) Database~\cite{seda:08} and Persian Emotional Speech Database (Persian ESD)~\cite{kesh:15}.

Persian ESD~\cite{kesh:15} is a semi-natural database which includes 470 utterances in five basic emotions including \emph{anger}, \emph{disgust}, \emph{fear}, \emph{happiness}, \emph{sadness}, as well as neutral state. For collecting this database, 90 sentences were evaluated by a large group of native Persian-speakers to make sure that they were emotionally neutral. Two native speakers of Persian, a 50-year old man and a 49-year old woman, were asked to articulate the sentences in target emotions. The speakers were semi-professional actors who had participated in acting classes for a while. Prior to recording sessions, the speakers were asked to read a scenario and imagine experiencing the situation. Five scenarios (each corresponding to an emotional state) were used in this project. Thirty-four native speakers validated the database by recognizing the underlying emotion of each utterance in a 7-point nominal scale. According to the perceptual study, they achieved an accuracy of 71.4\% which is five times chance performance. 

SES~\cite{seda:08} is a simulated dataset which includes 1200 utterances or 50 minutes of speech data. To record SES, 10 university students (5 males and 5 females) were asked to read 10 single words, 12 sentences and 2 passages in four basic emotions of \emph{surprise}, \emph{happiness}, \emph{sadness}, \emph{anger} plus neutral mode. After recording the database, 24 annotators listened to the utterances only once and recognized the conveyed emotional state on a 5-point nominal scale. The annotators achieved 42.66\% accuracy in classifying emotions which was twice what would be expected by chance. 

Compared to the SES in terms of linguistic structure, the Persian ESD contains sentences with a single grammatical structure (subject + object + prepositional phrase + verb). The SES, however, covers various linguistic forms including word, sentence and passage. The questionnaire used in the SES perceptual study has a 5-point nominal scale which only includes the target emotions. Therefore, the participants were forced to choose an option from the given short list of emotions. Russell~\cite{rus:94} argues that not allowing listeners to label emotions freely results in forging agreement. As a solution, Frank and Stennett~\cite{fran:01} suggest adding \emph{none of the above} to the response option. As a result, a part of the recognition accuracy reported in the SES can be artifact because of excluding \emph{none of the above} option. Moreover, neither the SES nor the Persian ESD has provided standard phonetic transcriptions, 
so it is difficult to extract the linguistic content from their utterances. Table~\ref{tab:01} summarizes the Persian emotional speech databases in terms of accessibility, number of utterance, number of speakers, type of emotions, naturalness, pre-written text (scripted/unscripted), audio/visual mode and validation. 

\begin{table*}
\caption{Persian emotional speech databases. *ShEMO is added for the sake of comparison.}    
\label{tab:01}   
\rotatebox{90}{
\begin{tabularx}{\textheight}{m{0.97cm}m{1.5cm}m{1.4cm}m{2.95cm}m{2.95cm}p{3cm}m{0.91cm}m{0.91cm}m{1.1cm}} \hline
data & access & size & speaker & emotion & naturality & scripted & mode  & validation \\ \hline

\small{\cite{moos:07}} & \small{private} & \small{40} & \small{one with patois} & \small{anger, happiness, sadness, neutral mode} & \small{simulated (recorded in a non-acoustic environment and includes background noise)} 
 & \small{yes} & \small{audio}  & \small{not reported}  \\ \hline
 
\small{\cite{ghar:08}} & \small{private} & \small{116+1800 neutral utterances selected from Farsdat~\cite{bij:94}}& \small{one male}  & \small{sadness, anger, neutral mode}  & \small{simulated} & \small{yes} & \small{audio}  &  \small{conducted but not reported} \\ \hline

\small{\cite{seda:08}} & \small{commercially available} & \small{1200} & \small{5 females, 5 males}  & \small{sadness, happiness, surprise, anger, neutral mode} & \small{simulated} & \small{yes} & \small{audio} & \small{24 students evaluated the utterances} \\ \hline

\small{\cite{ghar:08}} & \small{private} & \small{252} & \small{22}  & \small{happiness, anger, interrogative, neutral mode} & \small{simulated} &\small{yes} &\small{audio} & \small{not reported}  \\ \hline

\small{\cite{man:09}} & \small{private}& \small{26} & \small{not reported} & \small{anger, fear, disgust, sadness, happiness, surprise} & \small{semi-natural (speakers were given pre-written scenarios \& asked to imagine the situation)} & \small{yes} & \small{audio \& visual}  & \small{not reported} \\ \hline

\small{\cite{ham:12}} & \small{private}& \small{2400}& \small{330 actors \& actresses} & \small{happiness, fear, sadness, anger, disgust, neutral mode} & \small{semi-natural (utterances were extracted from more than 60 Persian movies)} & \small{no}& \small{audio}  & \small{not reported} \\ \hline

\small{\cite{esm:13}} & \small{private} & \small{748} & \small{33 professional actors (18 males, 15 females)} & \small{anger, fear, sadness, happiness, boredom, disgust, surprise, neutral mode} & \small{semi-natural (collected from radio plays)} & \small{no}  & \small{audio} & \small{not reported}  \\ \hline

\small{\cite{kesh:15}} &  \small{public \& free} & \small{470} & \small{one actor, one actress}& \small{anger, disgust, fear, happiness, sadness, neutral mode} & \small{semi-natural (actors were given scripted scenarios for each emotion)} & \small{yes} & \small{audio} & \small{3 different evaluations were performed} \\ \hline 

\small{\cite{sav:15}} & \small{private} & \small{6720 (3-7 seconds each)} &  \small{10 males, 10 females}& \small{anger, fear, sadness, happiness, boredom, disgust, surprise, neutral mode} & \small{simulated}  & \small{yes} & \small{audio} &  \small{not reported} \\  \hline

\rowcolor[gray]{.95}\small{*ShEMO} & \small{public \& free} &\small{3000} & \small{31 females, 56 males}  & \small{anger, fear, happiness, sadness, surprise, neutral mode} & \small{semi-natural} & \small{yes}  & \small{audio} & \small{yes} \\ 
\noalign{\smallskip}\hline
        
\end{tabularx}}
\end{table*}

\section{Sharif Emotional Speech Database}
\label{sec:2}
Sharif Emotional Speech Database (ShEMO) is a large-scale semi-natural database for Persian which contains 3 hours and 25 minutes of speech data from 87 native-Persian speakers (31 females, 56 males). There are 3000 utterances in \emph{.wav} format, 16 bit, 44.1 kHz and mono which cover five basic emotions of \emph{anger}, \emph{fear}, \emph{happiness}, \emph{sadness} and \emph{surprise}, as well as neutral state. The utterances are extracted from radio plays which are broadcast online\footnote{www.radionamayesh.ir}. In the following subsections, we ellaborate different phases of developing ShEMO, including pre-processing, annotation and measuring reliability. 

\subsection{Pre-processing, Annotation and Reliability}
We selected $50$ radio plays of various genres including comedy, romantic, crime, thrilled and drama as potential sources of emotional speech. We balanced out the differences of the audio streams using a free open-source audio editor software application, named \emph{Audacity}. Since most streams (about 90\% of them) had a sampling frequency of 44.1kHz, we upsampled the streams which had a lower sampling rate using cubic interpolation technique. We also converted the stereo-recorded streams to mono. Mono channel is commonly used in speech communications where there is only one source of audio whereas stereo channel is usually applied when there are more than one source of audio. For example, using stereo channel in music production applications (where sound is generated from different instruments) leads to high-quality extraction and separation of multiple instruments from a single purely monophonic audio recording. Therefore, using stereo channel in speech applications only results in increasing bandwidth and storage space.

We segmented each stream into smaller parts such that each segment would cover the speech sample of only one speaker without any background noise or effect. We recruited 12 annotators (6 males, 6 females) to label the affective state of the utterances on a 7-point scale (including \emph{anger}, \emph{fear}, \emph{neutrality}, \emph{happiness}, \emph{sadness}, \emph{surprise}, and \emph{none of the above}). The annotators were all native speakers of Persian with no hearing impairment or psychological problems. The mean age of the annotators was 24.25 years (SD = 5.25 years), ranging from 17 to 33 years. Tables 2 and 3 highlight the detailed information of anonymous annotators.

\begin{table}[ht]
\centering
\caption{Detailed information of annotators}
\label{tab:02} 
\begin{tabularx}{\textwidth}{p{0.17\textwidth}p{0.17\textwidth}p{0.17\textwidth}p{0.5\textwidth}}

\hline\noalign{\smallskip}
 code &  gender &  age &  education \\ 
\noalign{\smallskip}\hline\noalign{\smallskip}
01 & male & $23$ & undergraduate student \\  
02 & female & $18$ & associate student \\
03 & female & $20$ & associate student \\ 
04 & male & $22$ & undergraduate student \\ 
05 & male & $31$ & PhD candidate \\ 
06 & male & $33$ & master degree \\ 
07 & female & $31$ & master degree \\ 
08 & male & $21$ & undergraduate student \\ 
09 & female & $23$ & undergraduate student \\ 
10 & female & $17$ & high school student \\ 
11 & male & $25$ & master degree \\ 
12 & female & $27$ & master degree \\ 
\noalign{\smallskip}\hline
\end{tabularx}
\end{table} 

\begin{table}[ht] \centering
\caption{Statistical values of annotators' age in year}\label{tab:03} 
\begin{tabularx} {\textwidth}{p{0.27\textwidth}p{0.27\textwidth}p{0.5\textwidth}}

\hline\noalign{\smallskip}
gender & mean & standard deviation\\  
\noalign{\smallskip}\hline\noalign{\smallskip}
female & $22.66$ & $4.99$ \\  
 male & $25.83$ & $5.46$ \\ 
total &$24.25$ & $5.25$ \\ 
\noalign{\smallskip}\hline
\end{tabularx}
\end{table} 

The utterances were randomly played in a quiet environment. The annotators were instructed to select \emph{none of the above} where more than one emotion was conveyed from an utterance or the underlying emotion was not among the specified emotional states.
Since the utterances were extracted from radio plays, there was no guarantee that the lexical contents would be emotionally neutral. Therefore, there might be some cases where the affective state of the speaker implied from their speech would be in a stark contrast with the lexical content of the utterance. To resolve this ambiguity and avoid any confusion, the annotators were intentionally asked to label the emotional state of utterances only based on the ways it had been portrayed in speech, regardless to the lexical contents. To make a final decision on the labels of the utterances, we considered majority voting~\cite{mov:09, mow:09, kar:10}. The utterances for which the majority voting decided \emph{none of the above} were discarded from the database as they probably reflected multiple emotions or an emotion which our database did not cover.

We calculated Cohen's kappa statistics~\cite{coh:60} as a measurement of inter-rater reliability\footnote{Cohen's kappa ranges generally from 0 to 1, where large numbers indicate higher reliability and values near zero suggest that agreement is attributable to chance alone.}.  According to the kappa statistics, there was 64\% agreement on the labels which means there is ``substantial agreement''~\cite{lan:77} among the annotators\footnote{As Landis and Koch~\cite{lan:77} explain, $0.61 < kappa < 0.80$ is interpreted as ``substantial agreement'' among the judges.}. We discarded the utterances for which a low reliability was reported. 

The mean length of utterances is 4.11 seconds (SD = 3.41), ranging from 0.35 to 33 seconds. The detailed information of utterances are illustrated in Table~\ref{tab:04}.

\vspace{-4mm}
\begin{table}[ht]\centering
\caption{Number and duration of utterances per each gender and affective state (SD = standard deviation)}\label{tab:04} 
\begin{tabularx}{\textwidth}{p{0.15\textwidth}p{0.09\textwidth}p{0.09\textwidth}p{0.07\textwidth}|p{0.09\textwidth}p{0.09\textwidth}p{0.09\textwidth}p{0.045\textwidth}}

\hline\noalign{\smallskip}
\multirow{2}{2cm}{affective state} & \multicolumn{3}{c}{number} & \multicolumn{4}{c}{duration (in second)} \\ \noalign{\smallskip}\cline{2-8}\noalign{\smallskip}
 & female & male & total & min & max & mean & SD \\  
\hline\noalign{\smallskip}
anger & $455$ & $604$ &  $1059$ & $0.44$ & $22.42$ & $3.61$ & $2.63$ \\  
fear & $22$ & $16$ & $38$ & $0.76$ & $8.97$  & $3.17$ & $1.84$\\
happiness & $111$ & $90$ & $201$  & $0.82$ & $13.39$ & $3.81$ & $2.36$ \\
neutral & $284$ & $744$ &  $1028$  & $0.56$ & $33.32$ & $4.89$ & $4.1$\\  
sadness & $271$ & $178$ &  $449$ & $0.69$ & $27.89$ & $4.84$ & $3.7$\\ 
surprise & $120$ & $105$ &  $225$  & $0.35$& $10.95$ & $1.79$ & $1.45$\\ 
total & $1263$ & $1737$ &  $3000$   & $0.35$ & $33.32$ & $4.11$ & $3.41$ \\ \hline
\end{tabularx}
\end{table} 

As shown in Table~\ref{tab:04}, \emph{anger} and \emph{fear} have the highest and lowest number of utterances in the database. For female and male speakers, the maximum number of utterances belongs to \emph{neutral} mode and \emph{anger}, respectively. Mean length of the utterances conveying \emph{surprise} (mean=$1.79$, SD=$1.45$) is remarkably shorter than other emotions (total mean=$4.11$, total SD=$3.41$). On the other hand, the highest duration has been reported for \emph{sadness} (mean=$4.84$, SD=$3.7$). It can be due to the fact that people usually use frequent silences and stops within their speech when conveying \emph{sadness}.
 
The ShEMO database is also orthographically and phonetically transcribed according to the International Phonetic Alphabet (IPA)\footnote{The IPA was devised by the International Phonetic Association as a standardized representation of the sounds of oral language.}, which can be useful for extracting linguistic features. A sample of orthographic and phonetic transcription, along with its English translation for \emph{anger} is illustrated in Fig. 1.

\begin{figure}[ht] \label{fig:01} \centering
\includegraphics[width=0.95\textwidth]{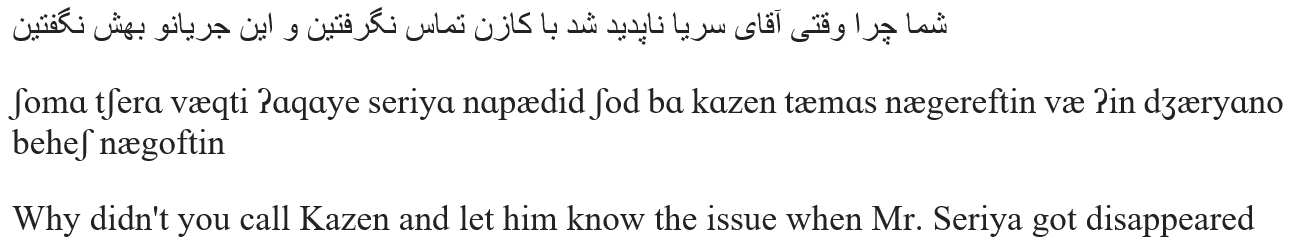}
\caption{Orthographic, phonetic and English translation of an utterance conveying anger}
\end{figure}

\subsection{Benchmark Results} \label{sec:3}
We provide baseline results of common classification methods on the ShEMO database. As features, we use the extended Geneva Minimalistic Acoustic Parameter Set (eGeMAPS)~\cite{eyb:16}. The eGeMAPS\footnote{It contains 88 different parameters. For further information, please refer to~\cite{eyb:16}.} refers to a basic standard acoustic parameter set including spectral (balance/shape/dynamics), frequency- and energy/amplitude-related parameters selected based on their potentiality for indexing affective physiological modifications in voice production, their automatic extractability, their proven value in previous studies and their theoretical importance~\cite{eyb:16}. We use the Munich Versatile and Fast Open-Source Audio Feature Extractor called openSMILE~\cite{eyben:10} to extract the eGeMAPS. To eliminate speaker variabilities, the features are normalized using z-score. 

We use three classifiers, namely support vector machine (SVM), \emph{k}-nearest neighbour (\emph{k}-NN), and decision tree (DT) for classification. According to Eyben et al.~\cite{eyb:16}, SVM is the most widely used static classifier in the field of speech emotion detection. Decision tree and \emph{k}-NN have also been applied for detecting the underlying emotional state of speech~\cite{grimm:07, lee:11}. We use SVM with Radial Basis Function (RBF) kernel and for decision tree model, we use random forest algorithm~\cite{bre:00} which are typical approaches in classification tasks. We apply nested cross-validation using one-vs-one multi-class strategy. The nested cross validation effectively uses a series of train/validation/test set splits to optimize each classifier's parameters and unbiasedly measure the generalization capabilities of classifier~\cite{caw:10}. In our work, first an inner 10-fold cross validation uses Bayesian optimization~\cite{sno:12} to tune the parameters of each classifier and select the best model. Then, an outer 5-fold cross validation is used to evaluate the model selected by the inner cross validation. Finally, we report the Unweighted Average Recall (UAR), which is popular in this field~\cite{schu:09, schu:10, met:11, schu:11, deng:12, schu:16}, averaged over the evaluation results for each classifier. Table~\ref{tab:05} demonstrates the performance of each classifier.  The range of parameters used in Bayesian optimization is given below: 

\begin{itemize}

\item \emph{k}-NN: Number of neighbours is set from $1$ to $30$,

\item \emph{k}-NN: Distance metrics include euclidean, cosine, chebychev and cubic.

\item SVM: Sigma (the kernel scale) and box (a cost to the misclassification) values are chosen between $0.00001$ and $100000$.

\item DT: Minimum observations number per leaf node is selected from $1$ to $20$. 

\item DT: Number of predictors (features) at each node is chosen from $1$ to the number of feature variables.

\end{itemize}

As presented in Table 5, SVM outperforms \emph{k}-NN and decision tree in both gender-dependent and -independent experiments. Decision tree has a better performance in comparison with \emph{k}-NN in gender-dependent experiment; however, it is slightly worse in gender-indepenent case. The best result is achieved for the SVM model trained on the female subsection of the data. Although the number of utterances is lower for female speakers (1263 vs. 1737 for male), they have a higher inter-rater reliability (kappa = 0.67 vs. 0.61 for male). In other words, the annotators had a stronger level of consensus on the underlying affective state of the female utterances. This may be the reason why SVM and \emph{k}-NN have better performance on the female subsection of the data. The confusion matrix of the performance in gender-independent mode (i.e. SVM = 58.2) is shown in Table 6. It should be mentioned that we excluded the fear utterances in our classification experiment because there was a small number of them in the database (38 in total). 

\vspace{-3mm}
\begin{table} [ht]\centering
\caption{Mean UAR obtained for SVM, \emph{k}-NN and decision tree using female, male and all utterances of the ShEMO}\label{tab:05} 
\begin{tabularx}{\textwidth}{p{0.17\textwidth}p{0.2\textwidth}p{0.2\textwidth}p{0.4\textwidth}} 
\hline\noalign{\smallskip}
 & SVM & \emph{k}-NN & DT\\
\hline\noalign{\smallskip}
female & $59.4$ & $47.4$ & $49.0$ \\ 
male & $57.6$ & $45.6$ & $46.6$ \\  
all & $58.2$ & $47.6$ & $47.4$ \\ \hline 
\end{tabularx}
\end{table} 

\vspace{-4mm}
\begin{table}[ht] \centering
\caption{Confusion matrix of the best performance in gender-independent mode}\label{tab:06} 
\begin{tabularx}{\textwidth}{p{0.18\textwidth}p{0.13\textwidth}p{0.13\textwidth}p{0.13\textwidth}p{0.13\textwidth}p{0.13\textwidth}}
\hline\noalign{\smallskip}
& anger & happiness & neutrality & sadness & surprise  \\
\noalign{\smallskip}\hline\noalign{\smallskip}
anger &  $\bf 911$ & $18$ & $85$ & $23$ & $22$ \\
happiness &  $\bf 68$ & $37$ & $59$ & $25$ & $12$ \\
neutrality & $69$ & $12$ &  $\bf 902$ & $33$ & $12$ \\
sadness & $37$ & $13$ & $107$ &  $\bf 263$ & $29$ \\
surprise & $31$ & $13$ & $56$ & $35$ &  $\bf 90$ \\ \hline
\end{tabularx}
\end{table} 

According to the confusion matrix, the model has the best performance in detecting \emph{anger} and \emph{neutrality}. The reason is that both \emph{anger} and \emph{neutral} mode have the highest number of utterances in the database, so the model properly learns the parameters associated with these two emotional states. On the other hand, the worst classification performance is reported for \emph{happiness} which has the lowest number of utterances in our data\footnote{\emph{Happiness} has the lowest number of utterances after \emph{fear}. As mentioned before, fear utterances were ignored in the classification experiments.}. According to Table 6, \emph{happiness} is mostly confused with \emph{anger}. \emph{Anger} and \emph{happiness} are categorized into high-arousal emotions; this can be the reason why the model has a poor performance in discriminating these two. As Scherer~\cite{sch:86} argues, emotions which are in the same category in terms of valence and arousal are usually confused with each other. Moreover, \emph{anger}, \emph{sadness} and \emph{surprise} are confused with \emph{neutrality}. It seems that this happens for the utterances with lower emotional strength. Moreover, the utterances conveying \emph{surprise} are relatively short; therefore, it can be challenging for the model to differentiate \emph{surprise} from other emotional states based on a short context. 

In order to compare our baseline model and see how it works on other databses and languages, we train and test the mentioned classifiers on Persian ESD~\cite{kesh:15}, Berlin Emotional Speech database (EMO-DB)~\cite{bur:05} and the Ryerson Audio-Visual Database of Emotional Speech and Song (RAVDESS)~\cite{liv:12}. The EMO-DB covers $535$ speech samples of $10$ speakers ($5$ males, $5$ females) in $6$ emotional states of \emph{anger}, \emph{boredom}, \emph{disgust}, \emph{anxiety/fear}, \emph{happiness} and \emph{sadness}, as well as neutral mode. The EMO-DB is a balanced, simulated database with pre-written texts\footnote{Actors were asked to read $10$ short emotionally neutral sentences.}. The RAVDESS database includes $1440$ utterances\footnote{We trained the models on the audio (not video), speech (not song) files of the database.} articulated by $24$ ($12$ males, $12$ females) speakers in a north American English accent. It includes $7$ emotional states: \emph{happiness}, \emph{sadness}, \emph{anger}, \emph{calmness}, \emph{fear}, \emph{surprise} and \emph{disgust}, as well as neutral mode. The RAVDESS is a balanced, simulated database and uses professional actors to record the utterances. Fig. 2 illustrates the results of comparison.

\begin{figure}[ht] \centering \label{fig:02}
\includegraphics[width=0.9\textwidth]{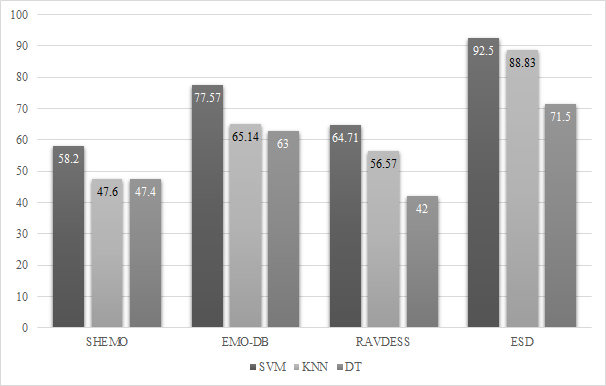}
\caption{Comparison of SVM, \emph{k}-NN and decision tree on the ShEMO (Persian), EMO-DB (German), RAVDESS (English) and Persian ESD datasets. The vertical axis indicates UAR.}
\end{figure}

As shown in Fig. 2, the classifiers trained on the Persian ESD result in the highest UAR. On the contrary, the decision tree trained on the RAVDESS dataset and SVM and \emph{k}-NN trained on the ShEMO have the lowest performance. All databases, except for the ShEMO, are balanced and have a fixed prompt for all speakers and emotions. On the other hand, the ShEMO is unbalanced and more realistic data, so detecting the underlying affective state of the utterances is a harder task in this case. As the results indicate the ShEMO would provide the research community with challenges in developing proper classification techniques for emotion detection in more realistic environments. 

\section{Summary, Conclusion and Future Work} \label{sec:4}
This paper introduces a large-scale validated dataset for Persian which contains semi-natural emotional, as well as neutral speech of a wide variety of native-Persian speakers. In addition to the database, we present the benchmark results of common classification methods to be shared among the researchers of this field. 

Our immediate future work includes increasing the number of utterances for fear. We also intend to extend the benchmark results to include other classification methods such as hidden Markov models and deep neural networks as the state-of-the-art technique in speech emotion detection.  Labelling the data in terms of arousal and valence is another potential future extension.  Moreover, it would be interesting to study the frequency of neutral and emotional speech among native-Persian speakers and see whether the distribution of utterances in the ShEMO conforms to the standard distribution of emotions in Persian. In future, we can also annotate the emotional strength of utterances.

\begin{acknowledgements}
We would like to thank the anonymous reviewers for their insightful comments and suggestions. We also gratefully thank Dr. Steve Cassidy for his helpful points.
\end{acknowledgements}

\bibliographystyle{apalike}    
\bibliography{mybib}

\end{document}